\documentclass[journal] 
{IEEEtran}

\newtheorem{theorem}{Theorem}

\newtheorem{definition}{Definition}
\newtheorem{corollary}{Corollary}

\usepackage[linesnumbered,ruled,]{algorithm2e}
\usepackage{amsmath}
\usepackage{amssymb}
\usepackage{amsfonts}
\usepackage{mathrsfs}
\usepackage{subfigure}
\usepackage{epsfig}
\usepackage{graphicx}
\usepackage{color}
\usepackage{multirow}

\usepackage{etoolbox}

\begin{document}
\title{
Aligning Partially Overlapping Point Sets: an Inner Approximation Algorithm
}

\author{Wei~Lian,
	Wangmeng Zuo
and        Lei~Zhang 
\IEEEcompsocitemizethanks{
\IEEEcompsocthanksitem W. Lian is with the Department of Computer Science, 
Changzhi University, Changzhi,  046011, Shanxi, China,
and  the Department of Computing,
The Hong Kong Polytechnic University, Hong Kong.
E-mail: \emph{lianwei3@gmail.com}.
\IEEEcompsocthanksitem W.-M. Zuo is with School of Computer Science and Technology, Harbin Institute of Technology (HIT), Harbin, China.  
E-mail: \emph{cswmzuo@gmail.com}.
\IEEEcompsocthanksitem L. Zhang is with the Department of Computing,
The Hong Kong Polytechnic University, Hung Hom, Kowloon, Hong Kong.
E-mail: \emph{cslzhang@comp.polyu.edu.hk}.

}
}

\IEEEcompsoctitleabstractindextext{%
\begin{abstract}
	Aligning partially overlapping point sets 
where there is no prior information about the value of the transformation
is a challenging problem in computer vision.
To 
achieve this goal,
we first  reduce the objective  of the robust point matching algorithm to a function of a low dimensional variable.
The resulting function, however, 
is  only concave over a finite region including the feasible region.
To cope with this issue,
we  employ the inner approximation optimization algorithm
which only operates within the region where  the   objective function  is concave.
Our algorithm does not need regularization on transformation, and thus can handle the situation
where there is no prior  information about the  values of the transformations.
Our method is also $\epsilon-$globally optimal and thus is guaranteed to be robust.
Moreover, its most computationally expensive  subroutine is a linear assignment problem which can be efficiently solved.
Experimental results demonstrate the better robustness  of the proposed method over state-of-the-art algorithms.
Our method is also  efficient when the number of transformation parameters is small.
\end{abstract}


}

\maketitle

\IEEEdisplaynotcompsoctitleabstractindextext
\IEEEpeerreviewmaketitle

\section{Introduction}
Point matching 
is a fundamental  problem
in  
computer vision, pattern recognition and medical image analysis.
Disturbances such as  deformation,  occlusion,  outliers and  noise 
often makes this problem challenging.
One way of achieving point matching  is through the optimization of the objective function of the robust point matching (RPM) algorithm \cite{RPM_TPS}. 
By eliminating the transformation variable,
the work of \cite{RPM_concave_PAMI} reduces the objective function of RPM to a concave quadratic function 
of point correspondence
with a low rank Hessian matrix.
It  then uses the normal rectangular algorithm, a variant of the branch-and-bound (BnB) algorithm, for optimization. 
%
But it requires that each model point has a counterpart in another point set, which limits the method's scope of applications.

To address this issue,
the work of \cite{RPM_model_occlude} reduces the objective function of RPM to a concave function of point correspondence, which, albeit  not quadratic, still has a low rank structure. 
\cite{RPM_model_occlude} then uses the normal simplex algorithm,
a variant of the BnB algorithm, for optimization.
However, 
\cite{RPM_model_occlude} requires that 
the objective function is concave over a set of  simplexes whose union includes the feasible region,
whereas the objective function is not necessarily concave outside the feasible region.
To address this issue,
it enlarges the concavity region of the objective function by adding 
a regularization on transformation where prior information about  values of the  transformation needs to be supplied.
Consequently, the method tends to generate transformations biased towards the prior  values. 
So the method may fail to  handle the situation 
where prior information about the values of the transformations is unknown.

To address this issue, in this paper, we propose an alternative concave optimization approach.
Instead of using the BnB algorithm which requires  concavity of the objective function over a sufficiently large region, we use the inner approximation algorithm \cite{book_concave_intro}
which only operates within the  region over which the  objective function is concave.
Thus, our method does not need regularization on transformation  and  is 
able to handle the situation that 
there is no prior information about the values of the transformation.
Our method is also $\epsilon-$globally optimal and thus is guaranteed to be robust.
Moreover, 
its  computationally expensive  subroutine is a linear assignment problem which can be efficiently solved.
%
\section{Related work}

\subsection{Heuristic based point matching  methods} 
The  methods  related to ours are those modeling  transformation and point correspondence.
ICP  \cite{ICP,ICP2}
iterates between estimating   point correspondence 
and updating  transformation.
But it is prone to be trapped in local minima 
due to   discrete nature of point correspondence.
RPM \cite{RPM_TPS} 
relaxes point correspondence to be fuzzily   valued and
uses deterministic annealing (DA) for optimization.
But  DA is biased towards matching the mass  centers of two point sets.
%
%

%
The second category of methods are those modeling only  transformation.
The CPD method \cite{CPD} casts  point matching  
as the problem of fitting a Gaussian Mixture Model (GMM)  representing one point set to another point set.
The GMMREG method  \cite{kernel_Gaussian_journal} uses GMMs to represent two point sets  
and minimizes the $L_2$ distance between them.
The Schr\"odinger distance transform  is used to represent point sets in \cite{SDT_match}
and  the  registration problem is converted into  that  of
computing the geodesic distance between two points on a unit Hilbert sphere.
Support vector parameterized Gaussian mixtures (SVGMs) has been  proposed in
\cite{mixture_SVC} to  represent point sets  using sparse  Gaussian components.
The efficiency of GMM based methods is improved by  using filtering 
to solve the  correspondence  problem 
in \cite{GMM_filtering}.



The above methods  are all heuristic schemes.
Therefore, they may not perform well 
when the point matching problem becomes difficult.

\subsection{Global optimization-based methods}

The branch-and-bound (BnB) algorithm is a popular global optimization technique.
It is used to align  3D shapes
based on the Lipschitz optimization theory  \cite{Lipschitz_3D_align}.
But the method assumes no occlusion or outliers. 
BnB is used to recover 3D rigid transformation in \cite{branch_bound_align}.
But the 
correspondence needs to be known a priori.
The Go-ICP method  \cite{Go-ICP} uses BnB  to optimize the ICP objective  
by exploiting the  structure of the geometry of 3D rigid motions. 
The fast rotation search (FRS) method \cite{BnB_consensus_rotate,BnB_consensus_project}
recovers  rotation between 3D point sets 
by using stereographic projections.
The general 6 degree rigid registration is accomplished by using a nested BnB algorithm.
The GOGMA method 
registers two point sets by aligning the GMMs constructed from the original point sets \cite{BnB_mixture_Gaussian}.
%
Liu \textit{et al.} proposed a
rotation-invariant feature  in   \cite{BnB_rigid_regist_decom}.
%
Straub \textit{et al.}
proposed 
a
novel way of tessellating  rotation space in \cite{BnB_Bayesian_mixture}.
%
The above methods  are all  targeted at rigid registration. 
Therefore, they  may not cope well with scaling  and non-rigid deformation.



\section{Reformulation of objective function 
}

In this section,
we mainly follow the work of \cite{RPM_model_occlude} to derive our objective function.
%
Suppose there are two point sets 
$\mathscr{X}=\{ \mathbf x_i,i=1,\ldots,n_x\}$
and
$\mathscr{Y}=\{ \mathbf y_j,j=1,\ldots,n_y\}$ 
in  $\mathbb R^{n_d}$  to be matched,
where the coordinates $\mathbf x_i=\begin{bmatrix}
x_i^1,\ldots,x_i^{n_d}
\end{bmatrix}^\top$
and $\mathbf y_j=\begin{bmatrix}
y_j^1,\ldots,y_j^{n_d}
\end{bmatrix}^\top$.
RPM  achieves point matching by essentially 
solving the following mixed linear assignment$-$least square problem:
\begin{align}
\min\ &E(\mathbf P,\boldsymbol{\varphi})=\sum_{i,j}p_{ij}\|\mathbf y_j- T(\mathbf x_i|\boldsymbol{\varphi})\|^2
\\
s.t.\ &\mathbf P  \mathbf 1_{n_y}\le  \mathbf 1_{n_x},\ 
\mathbf 1_{n_x}^\top \mathbf P\le  \mathbf 1_{n_y}^\top,\
\mathbf 1_{n_x}^\top \mathbf P  \mathbf 1_{n_y}=n_p,\  
\mathbf P\ge 0 
\label{k_card_P_const}
\end{align}
where $T(\mathbf x_i|{\boldsymbol{\varphi}})$ denotes a transformation with parameters $\boldsymbol{\varphi}$.
$\mathbf P$ denotes
a correspondence matrix
with element $p_{ij}=1$ indicating that there is a matching between $\mathbf x_i$ and $\mathbf y_j$ and $p_{ij}=0$ otherwise.
%
$\mathbf 1_{n_x}^\top \mathbf P  \mathbf 1_{n_y}=n_p$ means the number of matches is equal to   $n_p$,  a preset  positive integer.

Under the condition 
that  $T(\mathbf x_i|{\boldsymbol{\varphi}})$ 
is linear w.r.t. its parameters ${\boldsymbol{\varphi}}$,
i.e., $T(\mathbf x_i|{\boldsymbol{\varphi}})=\mathbf{J}(\mathbf x_i){\boldsymbol{\varphi}}$, 
where $\mathbf{J}(\mathbf x_i)$ is called the Jacobian matrix,
after eliminating $\boldsymbol{\varphi}$
via solving $\frac{\partial E}{\partial \boldsymbol{\varphi}}=0$ for $\boldsymbol{\varphi}$ and making substitution,
the objective function  can be written as:
\begin{align}
&E(\mathbf{P}) =   \mathbf 1_{n_x}^\top  \mathbf{P} \widetilde{\mathbf y}     
\notag\\
&-  \mathbf y^\top  (\mathbf{P}\otimes I_{n_d})^\top  \mathbf{J}  
(\mathbf{J}^\top  (\text{diag}(\mathbf{P}  \mathbf 1_{n_y}) \otimes I_{n_d}) \mathbf{J})^{-1} 
\mathbf{J}^\top  (\mathbf{P}\otimes I_{n_d})\mathbf y  
\end{align}
where the matrix  
$\mathbf J\triangleq\begin{bmatrix}
\mathbf J^\top(\mathbf x_1), \dots, \mathbf J^\top(\mathbf x_{n_x})
\end{bmatrix}^\top$ and
the vectors
$\mathbf y\triangleq\begin{bmatrix}
\mathbf y_1^\top, \dots, \mathbf y_{n_y}^\top
\end{bmatrix}^\top$,
$\mathbf {\widetilde y}\triangleq \begin{bmatrix}
\|\mathbf y_1\|_2^2, \dots, \|\mathbf y_{n_y}\|_2^2
\end{bmatrix}^\top$.
$\mathbf 1_{n_x}$ denotes the $n_x$-dimensional  vector of  all ones and
$\mathbf I_{n_d}$ denotes the $n_d-$dimensional identity matrix.

To facilitate optimization of $E$,
$P$ needs  to be vectorized.
We define the vectorization of a matrix as the concatenation of its rows,
denoted by $\text{vec}(\cdot)$. 
Let $\mathbf p\triangleq \text{vec}(\mathbf P)$. 
To obtain a  concise form of $E$,
we need to introduce  new denotations.
Let 
\begin{gather}
\mathbf 1_{n_x}^\top  P \widetilde{\mathbf y}=\boldsymbol\rho^\top \mathbf p,
\quad
\mathbf{J}^T (P\otimes I_{n_d})\mathbf y =\mathbf\Gamma\mathbf p,
\\
\text{vec}(  \mathbf{J}^T (\text{diag}(P  \mathbf 1_{n_y}) \otimes I_{n_d}) \mathbf{J} )=
\text{vec}( \mathbf{J}_2^T  ((P  \mathbf 1_{n_y}) \otimes I_{n_\varphi}) ) =\mathbf\Xi\mathbf p
\end{gather}
where $n_\varphi$ denotes the dimension of ${\boldsymbol{\varphi}}$ and
$
\mathbf{J}_2\triangleq\begin{bmatrix}
\mathbf{J}(\mathbf x_1)^T \mathbf{J}(\mathbf x_1),       \ldots,      \mathbf{J}(\mathbf x_{n_x})^T \mathbf{J}(\mathbf x_{n_x})
\end{bmatrix}^T
$.
Based on the fact that $
\text{vec}(\mathbf M_1\mathbf M_2\mathbf M_3)
= (\mathbf M_1\otimes \mathbf M_3^\top)\text{vec}(\mathbf M_2) 
$ for any multiplicable matrices $\mathbf M_1$, $\mathbf M_2$ and $\mathbf M_3$, 
we have
\begin{gather}
\boldsymbol\rho=\mathbf 1_{n_x}  \otimes\widetilde{\mathbf y},\
\mathbf \Gamma=(\mathbf{J}^\top \otimes\mathbf y^\top )\mathbf W^{n_x,n_y}_{n_d},\
\\ 
\mathbf \Xi=(\mathbf{J}_2^\top \otimes I_{n_\varphi}) \mathbf W^{n_x,1}_{n_\varphi} (I_{n_x} \otimes\mathbf 1_{n_y}^\top )
\end{gather}
Please refer to \cite{RPM_model_occlude} for definition of the constant matrix $\mathbf W^{n_x,n_y}_{n_d}$.

With the above preparation, 
$E$  can be rewritten in terms of vector $\mathbf p$ as:
\begin{gather}
E(\mathbf p)=
\boldsymbol \rho^\top \mathbf p
- \mathbf p^\top \mathbf\Gamma^\top  \text{mat} (\mathbf\Xi\mathbf p)^{-1} \mathbf\Gamma\mathbf p   \label{eng_vec}
\end{gather}
where $\text{mat}(\cdot)$
denotes reconstructing a symmetric matrix 
from a vector which is the  result of applying $\text{vec}(\cdot)$ to a symmetric matrix.
Thus, $\text{mat}(\cdot)$ can be viewed as the inverse of the operator $\text{vec}(\cdot)$.

Since $\mathbf  1^\top_{n_xn_y} \mathbf  p=n_p$, a constant,
rows in $\mathbf \Xi$
equal to scaled versions  of $\mathbf  1^\top_{n_xn_y}$ will be useless and can be discarded.
Also, redundant rows can  be removed.
Since $\text{mat}(\mathbf \Xi\mathbf p)$ is a symmetric matrix,
$\mathbf \Xi$    contains redundant rows.
Based on this  analysis,
we hereby  denote $\mathbf \Xi_2$ 
as the matrix formed as a result of $\mathbf \Xi$ 
removing such  rows.
Please refer to Section \ref{sec:exp} for examples of $\mathbf \Xi_2$.
%
In view of the form of $E$, 
we can see  that  $E$ is   determined by the variable
$\begin{bmatrix}
\mathbf \Xi_2^\top ,  \mathbf \Gamma^\top, \boldsymbol \rho\end{bmatrix}^\top\mathbf p = \mathbf R^\top \mathbf Q^\top\mathbf p$, 
which in turn is determined by a low dimensional variable $\mathbf u'\triangleq  \mathbf Q^\top\mathbf p$.
Here 
$
\mathbf Q\mathbf R
$ denotes the QR factorization of 
$
\begin{bmatrix}
\mathbf  \Xi_2^\top, \mathbf \Gamma^\top, \boldsymbol \rho\end{bmatrix}
$
with  $\mathbf R$ being an upper triangular matrix 
and the columns of $\mathbf Q$ being orthogonal unity vectors.
%
%
The specific form of $E$ in terms of variable ${\mathbf{u}'}$ is:
\begin{equation}
E(\mathbf{u}')= (\mathbf{R}^T \mathbf{u}')_{\rho}    - (\mathbf{u}'^\top \mathbf{R})_{\Gamma}   
[\text{mat} ((\mathbf{R}^\top \mathbf{u}')_{\Xi_2}) ]^{-1} (\mathbf{R}^\top \mathbf{u}')_{\Gamma} 
\end{equation}
where $(\mathbf R^\top\mathbf{u}')_{\Xi_2}$ 
denotes the vector formed by the elements of vector $\mathbf R^\top\mathbf{u}'$
with  indices  equal to row indices of the submatrix $\mathbf \Xi_2$ 
in matrix $\begin{bmatrix}
\mathbf \Xi_2^\top, \mathbf \Gamma ^\top,\boldsymbol \rho
\end{bmatrix}^\top$.
Vectors 
$(\mathbf R^\top\mathbf{u}')_{\Gamma}$ and $(\mathbf R^\top\mathbf{u}')_{\rho}$ are similarly defined.                                                                   
Here we abuse the use of 'mat' such that 
$\text{mat}(\mathbf \Xi_2\mathbf p) =\text{mat}(\mathbf \Xi\mathbf p)$.
The meaning will be clear from the context.

\section{Optimization}
The  analysis in the previous section indicates  that 
$E$ is  a function of the low dimensional variable $\mathbf{u}'$
with the  feasible region $U'\triangleq\{\mathbf{u}':\mathbf{u}'=Q^\top \mathbf p, \mathbf p\in \Omega\}$,
where $\Omega$ denotes the feasible region of $\mathbf p$,
as is defined by \eqref{k_card_P_const}.

%
Based on Proposition 1 in \cite{RPM_model_occlude},
one can see that 
$E$ is  concave over the spectrahedra
$\Psi' \triangleq\{\mathbf u' : 
\text{mat} ((\mathbf{R}^\top \mathbf{u}')_{\Xi_2})\succ0
\}\supset U'$.
Thus, it is natural to use the inner approximation  algorithm \cite{book_concave_intro},
a global optimization algorithm specifically designed for  functions which are concave  over a finite region, to optimize $E$.

\subsection{Translation of the coordinate system
	\label{sec:coord}
}


To facilitate further derivation,
it is convenient to  work in   a new coordinate system
which is constructed as follows.

We first solve 
a series of 
linear assignment problems
\begin{equation}
\max \{ \mathbf h_i^\top  \mathbf{u}':  \mathbf{u}'\in U' \},
\end{equation}
to obtain $n_u+1$  solutions  $\mathbf{v}'_i\in U'$. 
Here
$n_u$ denotes the dimension of $\mathbf u'$
and $\mathbf h_i, i=1,\ldots, n_u+1$ are preset
$n_u$-dimensional vectors
such that $\mathbf h_i- \mathbf h_{n+1}$ are linearly independent.
%
Different choices of  $\mathbf h_i$ are possible.
For simplicity, in this paper,
we choose  $\mathbf h_i$  as $\mathbf e_i$, $i=1,\ldots,n_u$ and $-\mathbf 1_{n_u}$, respectively.
Here $\mathbf e_i$ denotes the $n_u-$dimensional vector with the $i$-{th} element being $1$ and remaining elements being $0$s.
%
Let $\mathbf v'_0 =\frac{1}{n_u}\sum_i \mathbf{v}'_i $.
Apparently, $\mathbf v'_0\in \text{int} S'$,
where "$\text{int}$" denotes the interior of a convex set 
and the simplex
$S'\triangleq \text{co}\{\mathbf v'_1,\ldots,\mathbf v'_{n_u+1}\}\subset U'$.
Here $\text{co}\{\}$ denotes the convex hull of a point set.

Now we define the new coordinate system
as the result of  translating  the  coordinate system  of $\mathbf u'$ such that 
$\mathbf v'_0$ is the new origin.
Points $\mathbf{u}$ and $\mathbf{u}'$  in the new and old  coordinate systems are related by  $\mathbf{u}'=\mathbf u+ \mathbf v'_0$.
Accordingly, 
the energy function for $\mathbf{u}$ is $E_2(\mathbf{u}) \triangleq E(\mathbf{u}+\mathbf v_0)=E(\mathbf u')$.
Besides,
the feasible region of $\mathbf{u}$ is $ U\triangleq\{\mathbf{u}:\mathbf u=\mathbf Q^\top \mathbf p-\mathbf v'_0, \mathbf p\in \Omega\}$ and 
$E_2(\mathbf{u})$ is  concave over the spectrahedra
$\Psi \triangleq\{\mathbf{u}:   
\text{mat} ((\mathbf{R}^\top (\mathbf{u}+\mathbf v'_0))_{\Xi_2})\succ0
\}$.
%
%
Let the simplex
$S\triangleq \text{co}\{\mathbf v_1,\ldots,\mathbf v_{n_u+1}\}$, 
where the vertices $\mathbf v_i =\mathbf v'_i - \mathbf v'_0$.

It is noted   that, 
instead of using  a vertex of the feasible region $U'$ as the center of the new coordinate system
in \cite{book_concave_intro},
we use an interior point of $U'$ as the center of the new coordinate system.
This brings the benefit that 
the facet enumeration procedure as will be presented in Section \ref{sec_facet_enum}
can be simplified.

\subsection{The inner approximation algorithm
	\label{sec_inner_approx}
}

The basic idea of the inner approximation algorithm applied to  our problem is as follows:

Construct a sequence of polytopes (i.e., bounded polyhedrons) $D_1, D_2, \ldots$ such that
\begin{enumerate}
	\item 
	$\emptyset \neq D_k \cap U\subset  D_{k+1} \cap U$
	and
	$D_k\subset\Psi$ for $k\ge 1$.
	\item 
	
	an optimal solution $\boldsymbol \omega_1$ of $\min\{E_2(\mathbf u): \mathbf u\in D_1\cap U \}$ is available.
	\item 
	an optimal solution $\boldsymbol \omega_{k+1}$ of $\min\{E_2(\mathbf u): \mathbf u\in D_{k+1}\cap U \}$
	can be derived from an optimal solution $\boldsymbol \omega_k$ of $\min\{E_2(\mathbf u): \mathbf u\in D_k\cap U \}$.
\end{enumerate}
The procedure stops when $D_k\supseteq U$,
since, in this case, $\boldsymbol \omega_k$ is 
an optimal solution of 
$
\min\{ E_2(\mathbf u) : \mathbf u\in U \}
$.
The sequence $U\cap D_1,U\cap D_2,\ldots$
constitutes an inner approximation of $U$ by "expanding" polytopes.
The polytope $D_{k+1}$ can  be constructed from $D_k$ by choosing a suitable point $\widetilde{\mathbf z}_k\notin D_k$ and setting 
\begin{equation}
D_{k+1}=\text{co}\{D_k\cup \{\widetilde{\mathbf z}_k\}\}
\label{D_recursion}
\end{equation}

To ensure  convergence of the algorithm in finite iterations,
we require that $D_{k+1}\setminus D_k$ contains a vertex  of $U$.
Therefore, in each iteration,
the algorithm finds
a vertex $\mathbf z$ of $U$ satisfying  $\mathbf z\notin D_k$,
and determines $\widetilde{\mathbf z}_k$ in 
\eqref{D_recursion} from $\mathbf z_k$
such that 
\begin{equation}
D_{k+1}=\text{co} (D_k \cup \{\widetilde{\mathbf z}_k\}) \supseteq \text{co}(D_k \cup \{\mathbf z_k\})
\end{equation}
Usually $\widetilde{\mathbf z}_k \neq \mathbf z_k$,
thus, $D_{k+1}$ will be strictly larger than $\text{co}(D_k \cup \{\mathbf z_k\})$.
The purpose of using $\widetilde{\mathbf z}_k$ instead of $\mathbf z_k$ is to make $D_{k+1}$ as large as possible so as to improve  the convergence of the algorithm.
%
%
%
%
%
%
%
%


\subsection{Initial polytope }

Although  the simplex $S= \text{co}\{\mathbf v_1,\ldots,\mathbf v_{n_u+1}\}$  in Section \ref{sec:coord} can be used as the initial polytope,
it is advantageous that the initial polytope is chosen as large as possible so as  to improve the convergence  of our algorithm.
To this end,
We next expand  $S$ by using a simplified version of the
$\gamma-$extension \cite{book_concave_intro}
where we only specify directions.

\begin{definition}
	A point $\widetilde{\mathbf d}$ is called $\gamma-$extension in direction $\mathbf d\in \mathbb R^{n_u}\setminus \{0\}$ if
	\[
	\widetilde{\mathbf d}=\theta \mathbf d \quad \text{with} \quad \theta=
	\max\{t: E_2(t\mathbf d)\ge\gamma, t \mathbf d  \in \Psi\}
	\]
\end{definition}

We solve this problem by first solving the subproblem 
\begin{gather}
\max \{t:  t \mathbf d \in \Psi\}
\end{gather}
This  is a semidefinite program, 
for which solvers such as Sedumi \cite{SeDuMi} can be employed.
Suppose the optimal $t$ is $t_0$,
then we can use, e.g., the bisection algorithm to solve the second subproblem:
\begin{equation}
\max \{t: E_2(t\mathbf d)\ge \gamma,  0\le t\le t_0 \}
\end{equation}

Let
$
\widetilde{\mathbf v}_i = \theta_i \mathbf v_i
$
be the $\gamma-$extension  in direction $\mathbf v_i$ with  $\gamma=E_2(\boldsymbol \omega)$.
Here 
$\boldsymbol \omega =\arg\min_i E_2(\mathbf v_i)$ is the initial optimal solution.
Due to concavity of $E_2$ over $\Psi$, it follows that $\theta_i\ge 1$.
%
The unique hyperplane 
passing through $\widetilde{\mathbf v}_i, i=1,\ldots,j-1,j+1,\ldots,n_u+1$ is 
\begin{equation}
H^j=\{\mathbf u: \mathbf u=\mathbf Y_j \boldsymbol \lambda , \mathbf 1_{n_u}^\top \boldsymbol \lambda=1\}
=\{\mathbf u: \mathbf 1_{n_u}^\top\mathbf Y_j^{-1}\mathbf u=1 \}
\end{equation}
where the matrix $\mathbf Y_j=\begin{bmatrix}
\widetilde{\mathbf v}_1 & \ldots & \widetilde{\mathbf v}_{j-1}& \widetilde{\mathbf  v}_{j+1} &\ldots&\widetilde{\mathbf v}_{n_u+1}
\end{bmatrix} $.
Since the origin 
$\mathbf 0\in \text{int} S$,
it follows that
${\mathbf v}_1, \ldots, {\mathbf v}_{j-1},  {\mathbf v}_{j+1}, \ldots, {\mathbf v}_{n_u+1}$ 
are linearly dependent,
so do
$ \widetilde{\mathbf v}_1, \ldots,$ $\widetilde{\mathbf v}_{j-1},  \widetilde{\mathbf v}_{j+1}, \ldots,\widetilde{\mathbf v}_{n_u+1}$.
Thus, $\mathbf Y_j$ is invertible.
We define the half space  
$H^j_-\triangleq \{\mathbf u: 1_{n_u}^\top \mathbf Y_j^{-1} \mathbf u \le 1 \}$.
Apparently,  $\mathbf 0\in \text{int} H^j_-$.
We now set the initial polytope as the simplex
\begin{equation}
D=\cap_j H^j_{-}
\end{equation}
Thus,  $\mathbf 0\in \text{int} D$.

\subsection{Updating polytope}
At some stage of the algorithm, we have 
\begin{equation}
D=\{\mathbf u: \mathbf d_i^\top \mathbf u \le 1, i\in I  \}
\label{formu_D_ineq}
\end{equation}
with some finite index set $I$ and $\widetilde{\mathbf z}\in \mathbb R^{n_u}\setminus D$.
Then the next polytope
\begin{equation}
\overline{D}=\text{co}(D \cup \{\widetilde{\mathbf z}\})
\end{equation}
is of the form
\begin{equation}
\overline{D}=\{\mathbf u: \overline{\mathbf d}_i^\top \mathbf u \le 1, i\in \overline{I}\}
\end{equation}
Finding $\overline{\mathbf d}_i$ is a classical  facet enumeration problem
which will be treated in Section \ref{sec_facet_enum}.

\subsection{
	Termination condition
}
As shown in Section \ref{sec_inner_approx}, the algorithm will terminate if 
$U\setminus D=\emptyset$.
Since $D$ is of the form \eqref{formu_D_ineq},
we can check whether $U\setminus D=\emptyset$ by  solving the following linear assignment programs 
\begin{equation}
\mu_i=\max \{\mathbf d_i^\top \mathbf u : \mathbf u\in U \} \quad (i\in I)
\end{equation}
Then $Q=\{\mathbf u: \mathbf d_i^T \mathbf u\le \mu_i, i\in I\}$ is a polytope containing $U$, 
and we have $U\setminus D=\emptyset$ if and only if $D\supseteq Q$, i.e., if and only if  $\mu_i\le 1$ for each $i\in I$.

In this paper, instead of 
using the termination condition  $\max_i \mu_i\le 1$,
which is generally computationally  expensive,
we set the termination criterion  as  $\max_j \mu_j$ $\le 1+\epsilon$,
where 
$\epsilon$ is a preset small positive value.
Consequently, our algorithm becomes an $\epsilon-$globally optimal algorithm.
Since higher dimensional  space of $\mathbf u$ tends to  lead to slower convergence,
instead of directly setting $\epsilon$,
we let $\epsilon=n_u \epsilon_0$
by also taking into account  the dimension $n_u$ of the space of $\mathbf u$ and set $\epsilon_0$ instead.


\subsection{
	Expanding polytope
}
If $\max_i \mu_i > 1+\epsilon$,
then we need to expand the polytope $D$.
For  $j^*= \arg\max_i \mu_i$,
we have an optimal vertex solution 
$\mathbf z\in V(U)\setminus D$
of $\max\{\mathbf d_{j^*}^T \mathbf u: \mathbf u\in U  \}$.
Here $V(\cdot)$ denotes the vertex set of a polytope.



As is shown before,
a larger $D$  benefits the convergence of our algorithm.
To this end,
we choose $\widetilde{\mathbf  z}=\theta \mathbf z$ as  the $\gamma$-extension in direction $\mathbf z$ with $\gamma=\min\{E_2(\mathbf z), E_2(\boldsymbol \omega)\}$.
Due to concavity of $E_2$ over $\Psi$,
we have  $\theta\ge 1$.
We now set $\overline{D}=\text{co}(D\cup \{\widetilde{\mathbf z}\})$.
Meanwhile,
the optimal solution 
so far obtained
is updated as  $\boldsymbol \omega{\leftarrow} {\arg\min  \{E_2(\mathbf z), E_2(\boldsymbol \omega)\}}$.


\subsection{Facet enumeration
	\label{sec_facet_enum}
}

\textbf{Facet enumeration (FE) problem}:  Given a polytope $D$ of the form \eqref{formu_D_ineq}
and given a point $\widetilde{\mathbf z}\in \mathbb{R}^{n_u}\setminus D$,
problem FE aims to find the inequality representation of $\overline{D}=\text{co}(D\cup\{\widetilde{\mathbf z} \} )$.

Instead of directly solving problem FE which is challenging, 
following \cite{book_concave_intro},
we  use the concept of polars to equivalently transform this problem into the 
the vertex enumeration problem (VE)
and then solve the resulting problem.


\begin{definition}
	Let $D\subset R^{n_u}$ be a convex set. Then the set 
	\begin{equation}
	D^0=\{\mathbf v: \mathbf v^\top \mathbf u\le 1 \text{ for all }\mathbf u\in D  \}
	\end{equation}
	is called the polar of $D$.	
\end{definition}
Geometrically speaking,
the polar $D^0$ describes the set of normals to the hyperplanes $\{\mathbf u:\mathbf v^\top \mathbf u=1\}$ 
such that the  half spaces  $\{\mathbf u:\mathbf v^\top \mathbf u\le1\}$  contain $D$.
It is easy to see that $D^0$ is bounded if and only if the origin $\mathbf 0\in \text{int} D$.

\begin{theorem} \label{duality_v_f}
	(\textbf{vertex-facet duality}) 
	Let $D=\{\mathbf u: \mathbf d_i^\top\mathbf u\le 1, i\in I\}$, 
	$\mathbf d_i\in \mathbb{R}^{n_u}\setminus\{\mathbf 0\},i\in I$ be a polytope whose facets are defined by $\mathbf d_i$, and 	let
	$D^0$
	be the polar of $D$ with vertex set $V(D^0)$, Then 
	\begin{equation}
	V(D^0)=\{\mathbf d_i: i\in I\}
	\end{equation}
\end{theorem}
Please refer to \cite{book_concave_intro} for the proof.
Note  that compared with the corresponding  theorem  in  \cite{book_concave_intro},
Theorem \ref{duality_v_f} has a simpler form
without considering the extreme directions of $D^0$.
This is  because for our algorithm,
we have  $\mathbf 0\in \text{int} D$
and thus $D^0$ is bounded.

Returning to problem FE,
we see from  Theorem \ref{duality_v_f} that,
when switching to polars, 
this problem is equivalent to a vertex enumeration (VE) problem,
as is explained  in the following corollary:
\begin{corollary}
	Let $D^0$ and $\overline{D}^0=D^0\cap \{\mathbf u: \mathbf u^\top \widetilde{\mathbf z}\le 1 \}$ be the polars of $D$ and $\overline{D}=\text{co}(D\cup \{\widetilde{\mathbf z} \})$, respectively, 
	then each facet
	$\overline{D}\cap \{\mathbf u: \overline{\mathbf d}_i^\top \mathbf u=1\}$ of $\overline{D}$
	corresponds to a vertex $\overline{\mathbf d}_i$ of $\overline{D}^0$ and vice versa.
\end{corollary}

Given $V(D^0)$,
finding  $V(\overline{D}^0)$ is precisely the classical problem of VE \cite{book_concave_intro},
which will be reviewed  in the next section.

\subsection{Vertex enumeration}

\textbf{Vertex Enumeration (VE) Problem}: 
Let $D=\{\mathbf u: g_i(\mathbf u)=a_i^\top \mathbf u - b_i\le 0, i=1,\ldots,m\}$
be a polytope with known vertex set $V(D)$, and let 
$H=\{\mathbf u: g_{m+1}(\mathbf u)= a_{m+1}^\top\mathbf u - b_{m+1} = 0 \}$
be a hyperplane such that 
$ 
\overline{D}=D\cap H
$ 
is neither empty nor a facet of $D$.
Problem VE aims to determine the vertex set $V(\overline{D})$ of $\overline{D}$.

Let
\begin{gather}
V^{+}(D)=\{\mathbf v\in V(D): g_{m+1}(\mathbf v)>0  \}
\\
V^{-}(D)=\{\mathbf v\in V(D): g_{m+1}(\mathbf v)<0  \}
\end{gather}

Without loss of generality, 
we assume $|V^{-}| \le |V^{+}|$.
Here $|\cdot|$ denotes the cardinality of a set.
For each $\mathbf u\in V^-$,
denote by $\mathcal J(\mathbf u)$ the set of constraints of $D$ which are active at $\mathbf u$.
Because of the way $D$ is constructed,
vertex $\mathbf u$ is nondegenerate, 
thus, we have $|\mathcal J(\mathbf u)|=n_u$ and linear independence of the corresponding system of linear equations
\begin{equation}
a_i^\top  \mathbf u - b_i =0 \quad (i\in\mathcal J(\mathbf u))
\label{u_bounding_plane}
\end{equation}
Moreover,
$\mathbf u$ has 
$n_u$ neighboring vertices in $D$.
That is, 
$n_u$ edges of $D$ are incident with $\mathbf u$.
Each line through $\mathbf u$ in the direction of such an edge is the solution set of a system of $n_u-1$ linear equations
which can be obtained from \eqref{u_bounding_plane} 
by dropping one equation.
The set of new vertices in $D\cap H$ which are adjacent to $\mathbf u$ contains
the intersection points of these lines with the hyperplane $H$.

Without loss of generality,
for simplicity of notation,
we assume
$\mathcal J(\mathbf u)=\{1,\ldots$ $,n_u\}$.
Then, for each $\mathbf u\in V^-$,
we have to consider the $n_u$ systems of $n_u$ linear equations
\begin{align}
&a_i^\top \mathbf u - b_i =0 \quad (i\in\{1,\ldots,n_u\}\setminus \{l\})\notag\\
&a_{m+1}^\top \mathbf u - b_{m+1}=0
\label{edge_intersect_with_plane}
\end{align}
which arise when $l$ runs from $1$ to $n$.

When a system in \eqref{edge_intersect_with_plane} has a solution $\boldsymbol \omega$, 
we have to check whether $\boldsymbol \omega$ satisfies the remaining inequalities of $D$.

Instead of directly solving \eqref{edge_intersect_with_plane},
which is cumbersome,
in the following,
the simplex pivoting algorithm
is employed to solve this  problem.
It  works by introducing slack variables $\mathbf v\in R^{n_u}_+$ to write the binding inequalities of $J(\mathbf u)$ in the form
\begin{equation}
A \mathbf u + I_{n_u} \mathbf v =\mathbf b
\label{slack_binding_inequal_u}
\end{equation}
and the equation of $H$ in the form
\begin{equation}
a_{m+1}^\top \mathbf u + 0^\top\mathbf  v = b_{m+1}
\label{equ_hyperplane}
\end{equation}

One can  transform \eqref{slack_binding_inequal_u} into
\begin{equation}
I_{n_u} \mathbf u + A^{-1}\mathbf v =A^{-1}\mathbf b \\
\end{equation}
and transform \eqref{equ_hyperplane} (by adding to \eqref{equ_hyperplane} multiples of the rows of \eqref{slack_binding_inequal_u}) into
\begin{align}
&0^T \mathbf  u + \overline{a}^\top \mathbf v =\overline{b} \label{tran_equ_hyperplane}
\end{align}
from which all possible new vertices neighboring $\mathbf u$ can be obtained by pivoting on all the current nonbasic variables $\mathbf v$ in the row \eqref{tran_equ_hyperplane}.

\begin{figure}[t]
	\centering
	\newcommand{\scale}{0.14}
\includegraphics[width=1\linewidth]{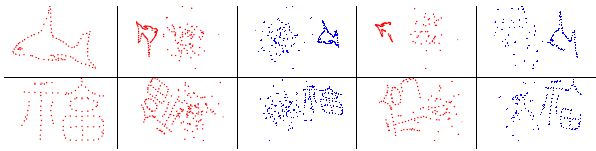}

	\caption{
		Left column: the prototype shapes.
		For the remaining columns:
		examples of model and scene point sets in the 
		outlier (columns 2, 3) and occlusion + outlier (columns 4, 5) tests. 
		\label{rot_2D_test_data_exa}}
\end{figure}

\begin{figure}[t]
	\centering
	\newcommand\scaleGd{0.27}
\includegraphics[width=1\linewidth]{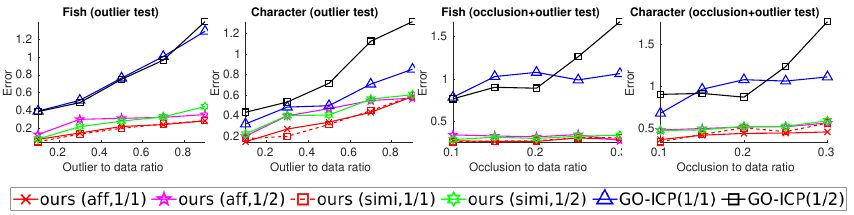}

	\caption{Average  matching errors by our method 
		and Go-ICP with different $n_p$ values (chosen from $1/2$ to $1/1$ the ground truth value)
		over 100 random trials
		for the 2D outlier and occlusion+outlier tests.
		\label{2D_simi_sta}}
\end{figure}

\begin{figure*}[t]
	\centering

 \includegraphics[width=\linewidth]{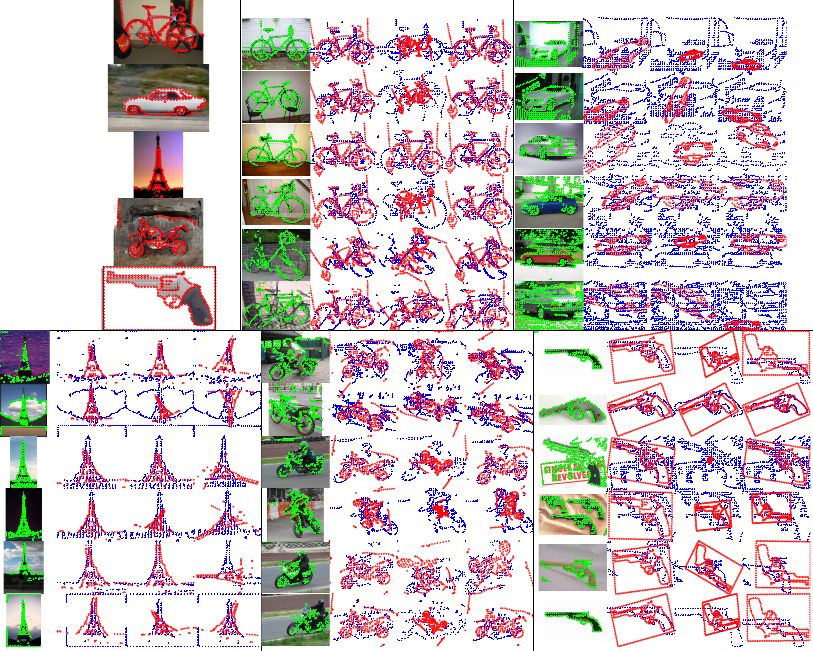}
	\caption{
		Top left grid cell: model  images with model point sets superimposed.
		The remaining  cells:
		scene images with scene point sets superimposed, alignment results by our method using similarity or affine transformations and Go-ICP.
		The $n_p$ value for every method is chosen as $0.9$ the minimum of the cardinalities of two point sets.
		\label{rot_2D_canny}}
\end{figure*}

\section{Experiments \label{sec:exp}}

We implement our method under  Matlab 2019b 
and compare it with other methods  on a PC with 3 GHz CPU and 32G RAM.
For the competing methods which only output point correspondences,
the generated correspondences are used to find the best affine transformations between  two point sets.
We define error  as  the root mean  squared difference
between the coordinates of  transformed ground truth model inliers and those of their corresponding scene inliers.
For our algorithm,
we set the parameter   $\epsilon_0=0.3$.

\subsection{2D synthesized datasets \label{subsec:regu2Dtest}}

We compare our method  with Go-ICP \cite{Go-ICP},
a globally optimal point set registration algorithm.
Go-ICP can handle  partial overlapping point sets and allows arbitrary rotation and translation
between two point sets.

2D similarity  and affine transformations  are respectively considered for our method.
For the former,
we have the formulation of the  transformation
\begin{equation}
T( \mathbf x_i| \boldsymbol\varphi)= \begin{bmatrix}
\varphi_1 x_i^1  - \varphi_2 x_i^2 +\varphi_3, &
\varphi_2 x_i^1 +\varphi_1 x_i^2  +\varphi_4
\end{bmatrix}^\top
\end{equation}
where   ${\boldsymbol\varphi}=\begin{bmatrix} \varphi_1, \ldots , \varphi_4 \end{bmatrix}^\top$.
Then we  have the Jacobian matrix
$\mathbf J(\mathbf x_i)=\begin{bmatrix}x_i^1&- x_i^2&1&0\\ x_i^2&x_i^1&0&1\end{bmatrix}$.
It can be verified that the rows of   
$\mathbf \Xi_2=\mathbf \Xi([1,3,4],:)$ 
constitute the unique  rows of $\mathbf \Xi$ 
not equal to scaled versions of $\mathbf 1_{n_xn_y}^\top$.

For 2D affine transformation,
we have the formulation of the  transformation
\begin{equation}
T( \mathbf x_i| \boldsymbol\varphi)= \begin{bmatrix}
\varphi_1 x_i^1  + \varphi_2 x_i^2 +\varphi_5, &
\varphi_3 x_i^1 +\varphi_4 x_i^2  +\varphi_6
\end{bmatrix}^\top
\end{equation}
where
${\boldsymbol\varphi}=\begin{bmatrix} \varphi_1, \ldots, \varphi_6 \end{bmatrix}^\top$.
Then we  have 
$\mathbf J(\mathbf x_i)=\begin{bmatrix}
x_i^1& x_i^2&0&0&1&0\\
0&0&x_i^1& x_i^2&0&1
\end{bmatrix}$.
It can be verified that the rows of 
$\mathbf \Xi_2=\mathbf \Xi([1,2,5,8,11],:)$ 
constitute the unique rows of $\mathbf \Xi$ 
not equal to scaled versions of $\mathbf 1_{n_xn_y}^\top$.

Following \cite{RPM_model_occlude},
two categories of tests are conducted:
1) \textbf{Outlier test}
and
2) \textbf{Occlusion + Outlier test}.
Different from \cite{RPM_model_occlude}, 
disturbances of
random rotation and scaling within range $[0.5,1.5]$ are also added  
when generating the model   point sets.
%
%
Fig. \ref{rot_2D_test_data_exa} illustrates these tests
and the prototype shapes.

The  matching errors by different methods
are presented in Fig. \ref{2D_simi_sta}.
One can see that our method using either transformation performs  better than Go-ICP,
particularly in the occlusion+outlier test,
where there is a large margin between the errors of our method and that  of Go-ICP.
In terms of different choices of  transformations,
our method using similarity or affine transformation performs  similar to each other. 
In terms of different choices of $n_p$,
our method with $n_p$  close to the ground truth  performs only slightly better.
This demonstrates  that our method is insensitive to different choices of  $n_p$.

The average running times (in seconds) by different methods are:
8.45 or
467.46   
for our method using similarity or affine  transformation
and 12.19 for Go-ICP.
This demonstrates  high efficiency of our method using similarity transformation.
Our method using affine transformation is two orders of magnitude slower  than our method using similarity transformation.
This is because affine transformation has  larger number of parameters.

%

\subsection{2D point sets extracted from images}
Point sets extracted from images are a more realistic setting for testing algorithms.
We test different methods on 2D
point sets extracted via the Canny edge detector  
from several images in  the Caltech-256 \cite{caltech_database} and VOC2007   \cite{pascal-voc-2007}
datasets, as illustrated in Fig. \ref{rot_2D_canny}.
To test a method's ability at handling  rotations,
model point sets are rotated 180 degree before being matched to scene point sets.

The registration results  by different methods
are presented in Fig. \ref{rot_2D_canny}.
%
One can see that our method using similarity transformation 
performs the best,
while Go-ICP and our method using affine transformation performs not well.
This is because affine transformation has 
more  transformation freedom than  rigid (which is used by Go-ICP) or similarity  (which is  used by another transformation version of our method)
transformation,
leading to the possibility of unconstrained registration results.
Another factor is that for our method,
the tolerance error $\epsilon=n_u\epsilon_0$ for affine transformation is actually larger than that of similarity transformation given that $\epsilon_0$ is set the same for both types of transformations.


\subsection{3D synthesized datasets without rotations \label{subsec:3Dtest_nonrot}}
Since 3D affine transformation contains  many parameters
which  causes    our method to converge too slowly,
it will not be tested. 
Instead,
%
We consider  the 3D transformation  consisting of nonuniform scaling  and translation for our method:
\begin{equation}
T( \mathbf x_i| \boldsymbol\varphi)= \begin{bmatrix}
\varphi_1 x_i^1  +\varphi_4, &   \varphi_2 x_i^2  +\varphi_5, &
\varphi_3  x_i^3  +\varphi_6
\end{bmatrix}^\top
\end{equation}
where $\boldsymbol\varphi=[\varphi_1,\ldots,\varphi_6]$.
We have the Jacobian matrix 
$\mathbf J(\mathbf x_i)=\begin{bmatrix}
x_i^1 & 0 & 0 & 1 & 0 & 0\\
0 &  x_i^2 & 0 & 0 & 1 & 0\\
0 & 0 &  x_i^3  & 0 & 0 & 1
\end{bmatrix}
$.
It can be verified that the rows of 
$\mathbf \Xi_2=\mathbf \Xi([1,4,8,11,15,18],:)$  
constitute the unique rows of $\mathbf \Xi$ 
not equal to scaled versions of $\mathbf 1_{mn}^\top$.

We  compare  our method  with RPM-BnB \cite{RPM_model_occlude}, RPM \cite{RPM_TPS}, CPD \cite{CPD} and GMMREG
\cite{kernel_Gaussian_journal}.
These methods 
only utilize the point position information for matching, and are capable of handling
partial overlapping point sets.
RPM-BnB is also globally optimal,
making it a good candidate for comparison.

\begin{figure}[t]
	\centering	
		\includegraphics[width=1\linewidth]{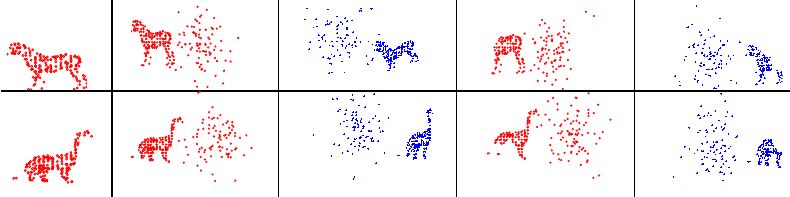}
 
	\caption{
		Left column: the prototype shapes. 
		For the remaining columns: 
		examples of model and scene point sets in the  outlier (columns 2, 3)
		and  occlusion+outlier (columns 4, 5) tests.
		\label{nonrot_3D_test_data_exa}}
\end{figure}

\begin{figure}[t]
	\centering
	\newcommand\scaleGd{0.27}
\includegraphics[width=1\linewidth]{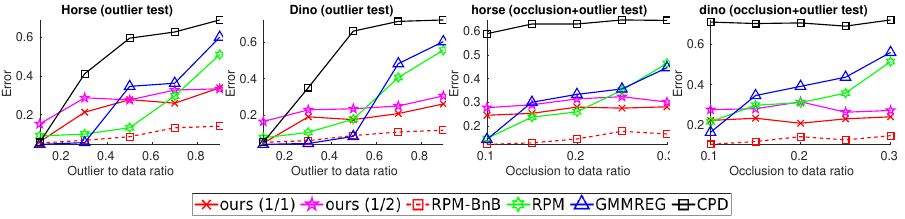}

	\caption{Average  matching errors by 		
		our method  with different $n_p$ values (chosen from $1/2$ to $1/1$ the ground truth value) and 
		other methods	over 100 random trials
		for the 3D outlier and occlusion+outlier tests.
		\label{3D_scal_tran_sta}}
\end{figure}

Analogous  to the experimental setup in Section \ref{subsec:regu2Dtest},
we conduct two categories  of tests:
1) \textbf{Outlier test} and
2) \textbf{Occlusion + Outlier test}.
Different from Section \ref{subsec:regu2Dtest},
no rotation disturbance is added 
when generating the  point sets. 
%
Fig. \ref{nonrot_3D_test_data_exa} illustrates these tests
and the prototype shapes.
The  matching errors by different methods  are presented in Fig. \ref{3D_scal_tran_sta}.
One can see  that 
our method performs slightly poorer than RPM-BnB.
Nevertheless, it is as robust as RPM-BnB
by performing  the same with 
increase of  severity of disturbance.
Note that our method is  more versatile (e.g., being able to handle 2D similarity invariant  alignment problem)
than RPM-BnB.
In comparison,
RPM, CPD and GMMREG only  perform
well when the disturbance is not severe.
The result also indicates that our method is relatively insensitive to different choices of $n_p$.



The average running times (in seconds) by different methods are: 
2769.5 for our method,
18.65 for RPM-BnB,
3.2 for RPM,
0.3 for GMMREG and
0.1 for CPD.

\subsection{3D synthesized datasets with rotations around  $z$-axis \label{subsec:3Dtest_rot}}

Next, we consider the 3D transformation  consisting of 
rotation around $z$-axis, uniform scaling on the $x$-$y$ plane, scaling along $z$-axis and translation 
for our method:
\begin{equation}
T( \mathbf x_i| \boldsymbol\varphi)= \begin{bmatrix}
\varphi_1 x_i^1  - \varphi_2 x_i^2 +\varphi_4, &
\varphi_2 x_i^1 +\varphi_1 x_i^2  +\varphi_5, &
\varphi_3  x_i^3  +\varphi_6
\end{bmatrix}^\top
\end{equation}
where $\boldsymbol\varphi=[\varphi_1,\ldots,\varphi_6]$.
We have the Jacobian matrix 
$\mathbf J(\mathbf x_i)=\begin{bmatrix}
x_i^1 & -x^2_i & 0 & 1 & 0 & 0\\
x_i^2 &  x_i^1 & 0 & 0 & 1 & 0\\
0 & 0 &  x_i^3  & 0 & 0 & 1
\end{bmatrix}
$.
It can be verified that the rows of 
$\mathbf \Xi_2=\mathbf \Xi([1,4,5,15,18],:)$  
constitute the unique rows of $\mathbf \Xi$ 
not equal to scaled versions of $\mathbf 1_{mn}^\top$.

\begin{figure}[t]
	\centering
	\newcommand\scale{0.16}		
\includegraphics[width=1\linewidth]{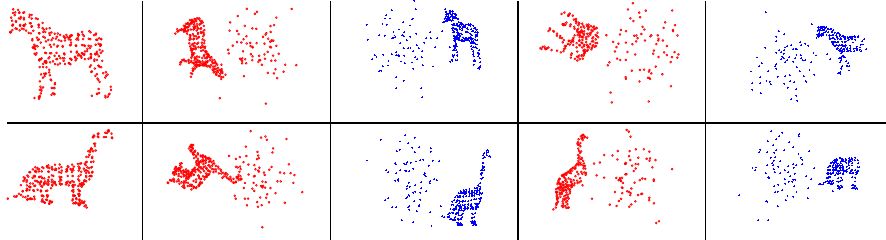}

	\caption{
		Left column: the prototype shapes.
		For the remaining columns:
		examples of model and scene point sets in the 
		outlier (columns 2, 3) and occlusion + outlier (columns 4, 5) tests. 
		\label{rot_3D_test_data_exa}}
\end{figure}

In this section,
besides Go-ICP, we also compare with FRS \cite{BnB_consensus_project}, 
which is based on global optimization,
only utilizes point coordinate information and 
allows arbitrary rotations and translations between two point sets.

\begin{figure}[t]
	\centering
	\newcommand\scaleGd{0.27}
\includegraphics[width=1\linewidth]{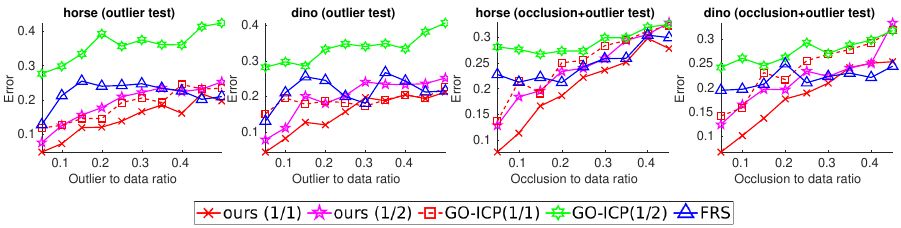}

	\caption{Average  matching errors by 		
		our method and Go-ICP with different $n_p$ values (chosen from $1/2$ to $1/1$ the ground truth value) and 
		FRS		over 100 random trials
		for the 3D outlier and occlusion+outlier tests.
		\label{3D_simi_sta}}
\end{figure}

Analogous  to the experimental setup in Section \ref{subsec:3Dtest_nonrot},
we conduct two categories  of tests:
1) \textbf{Outlier test} and
2) \textbf{Occlusion + Outlier test}.
Different from Section \ref{subsec:3Dtest_nonrot}, 
random rotation around the z-axis and uniform scaling within range $[0.5,1.5]$
is  applied to the prototype shape when generating the model point sets.
%
Fig. \ref{rot_3D_test_data_exa} illustrates these tests
and the prototype shapes.
The  matching errors by different methods are presented in Fig. \ref{3D_simi_sta}.
One can see that our method  performs overall better than other methods
and 
is  less  sensitive to different  choices of $n_p$ than Go-ICP.


The average running time (in seconds) by different methods are:
177.73  for our method,
69.26 for Go-ICP and
268.96  for FRS.

\section{Conclusion}
We proposed a global optimization-based algorithm for matching partially overlapping point sets.
%
It
works by reducing the RPM objective function to a function of a low dimensional  variable
and then using the  inner approximation algorithm to optimize the resulting objective function over its concave region.
%
Experiments on 2D and 3D data sets  demonstrated  better robustness of the proposed method over state-of-the-art algorithms for tasks involving  
various types of disturbances.
It  is also efficient when the number of transformation parameters is small. 

\subsection*{ACKNOWLEDGMENTS}
This work was supported by National Natural Science Foundation of China under Grant 61773002.

\let\mybibitem\bibitem
\renewcommand{\bibitem}[1]{%
	\ifstrequal{#1}{GM_IPFP}
	{\color{red}\mybibitem{#1}}
	{\color{black}\mybibitem{#1}}%
}

	\bibliographystyle{ieee}
	\bibliography{DP_SC_CVPR}

\end{document}